\documentclass[runningheads]{llncs}
\usepackage{graphicx}

\usepackage{tikz}
\usepackage{comment}
\usepackage{amsmath,amssymb} 
\usepackage{color}

\usepackage{booktabs}
\usepackage{epsfig}
\usepackage{multirow}
\usepackage{multicol}
\usepackage{enumerate}
\usepackage{array}
\usepackage[pagebackref,breaklinks,colorlinks]{hyperref}
\usepackage[capitalize]{cleveref}

\crefname{section}{Sec.}{Secs.}
\Crefname{section}{Section}{Sections}
\Crefname{table}{Table}{Tables}
\crefname{table}{Tab.}{Tabs.}
\newcommand{\mysubtitle}[1]{{\noindent}{\textbf{#1}}}

\usepackage[accsupp]{axessibility}  

\begin{document}
\pagestyle{headings}
\mainmatter
\def\ECCVSubNumber{5392}  

\title{Ghost-free High Dynamic Range Imaging with Context-aware Transformer} 

\titlerunning{Ghost-free High Dynamic Range Imaging with Context-aware Transformer}
%

\author{Zhen Liu$^{1,\ast}$ \and
Yinglong Wang$^{2,\ast}$ \and
Bing Zeng$^{3}$ \and
Shuaicheng Liu$^{3,1,\dagger}$}
\authorrunning{Z. Liu et al.}
%
\institute{Megvii Technology, Beijing, China \\
\email{liuzhen03@megvii.com} \\ \and
Noah's Ark Lab, Huawei Technologies, Shenzhen, China \\
\email{ylwanguestc@gmail.com} \\ \and
University of Electronic Science and Technology of China, Chengdu, China \\
\email{\{zengbing,liushuaicheng\}@uestc.edu.cn}\\
$^{*}$Joint First Author, $^\dagger$Corresponding Author} 

\maketitle

\begin{abstract}
High dynamic range (HDR) deghosting algorithms aim to generate ghost-free HDR images with realistic details. Restricted by the locality of the receptive field, existing CNN-based methods are typically prone to producing ghosting artifacts and intensity distortions in the presence of large motion and severe saturation. In this paper, we propose a novel Context-Aware Vision Transformer (CA-ViT) for ghost-free high dynamic range imaging. The CA-ViT is designed as a dual-branch architecture, which can jointly capture both global and local dependencies. Specifically, the global branch employs a window-based Transformer encoder to model long-range object movements and intensity variations to solve ghosting. For the local branch, we design a local context extractor (LCE) to capture short-range image features and use the channel attention mechanism to select informative local details across the extracted features to complement the global branch. By incorporating the CA-ViT as basic components, we further build the HDR-Transformer, a hierarchical network to reconstruct high-quality ghost-free HDR images. Extensive experiments on three benchmark datasets show that our approach outperforms state-of-the-art methods qualitatively and quantitatively with considerably reduced computational budgets. Codes are available at \url{https://github.com/megvii-research/HDR-Transformer}.
\end{abstract}

\begin{figure}[t]
  \centering
  \includegraphics[width=1.0\linewidth]{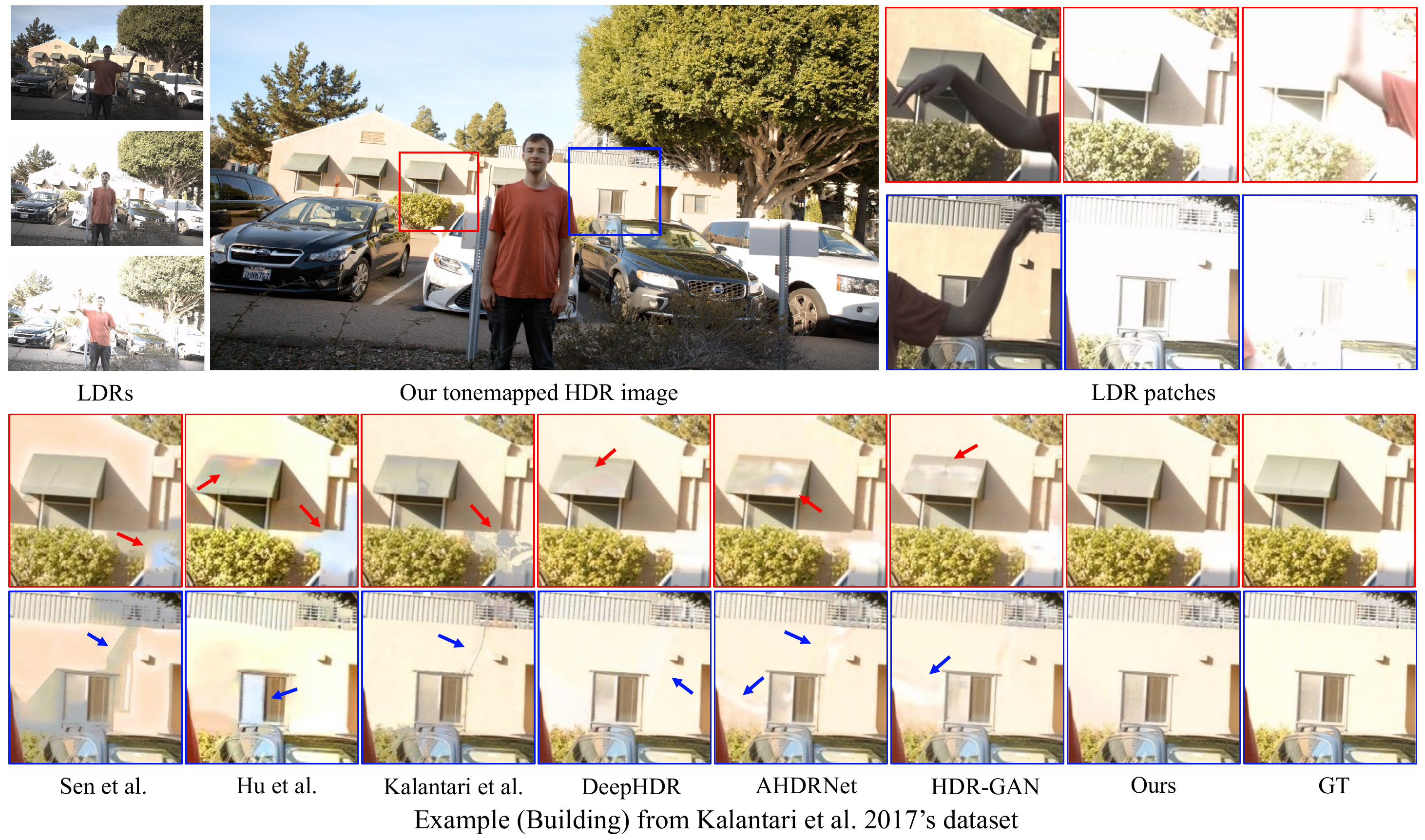}   
  \caption{Visual comparisons with the state-of-the-art methods~\cite{sen2012robust,hu2013hdr,kalantari2017deep,wu2018deep,yan2019attention,niu2021hdr} on Kalantari \emph{et al}~\cite{kalantari2017deep}’s dataset. As shown, the patch-match based methods~\cite{sen2012robust,hu2013hdr} and the CNN-based methods~\cite{kalantari2017deep,wu2018deep,yan2019attention,niu2021hdr} fail to remove the long-range ghosts caused by large motion and hallucinate reasonable local details in saturated regions. On the contrary, the proposed HDR-Transformer can effectively remove the ghosting artifacts and produce visual consistent local details.}\label{fig:teaser}
\end{figure}

\section{Introduction}

Multi-frame high dynamic range (HDR) imaging aims to generate images with a wider dynamic range and more realistic details by merging several low dynamic range (LDR) images with varying exposures, which can be well fused to an HDR image if they are aligned perfectly~\cite{ram2017deepfuse,raman2011reconstruction,ma2017robust,zhang2011gradient,mertens2007exposure,ma2019deep}. In practice, however, this ideal situation is often undermined by camera motions and foreground dynamic objects, yielding unfavorable \textit{ghosting artifacts} in the reconstructed HDR results. Various methods, commonly referred to as \textit{HDR deghosting algorithms}, have thus been proposed to acquire high-quality ghost-free HDR images.

Traditionally, several methods propose to remove ghosting artifacts by aligning the input LDR images~\cite{bogoni2000extending,hu2013hdr,kang2003high,zimmer2011freehand} or rejecting misaligned pixels~\cite{gallo2009artifact,grosch2006fast,pece2010bitmap,jacobs2008automatic,khan2006ghost} before the image fusion. However, accurate alignment is challenging, and the overall HDR effect is diminished when useful information is dropped by imprecise pixel rejection. Therefore, CNN-based learning algorithms have been introduced to solve ghosting artifact by exploring deep features in data-driven manners. 

Existing CNN-based deghosting methods can be mainly classified into two categories. In the first category, LDR images are pre-aligned using homography~\cite{hartley2003multiple} or optical flow~\cite{baker2011database}, and then multi-frame fusion and HDR reconstruction are performed using a CNN~\cite{kalantari2017deep,prabhakar2019fast,prabhakar2020towards,wu2018deep}. However, homography cannot align dynamic objects in the foreground, and optical flow is unreliable in the presence of occlusions and saturations. Hence, the second category proposes end-to-end networks with implicit alignment modules~\cite{yan2019attention,liu2021adnet,chung2022high} or novel learning strategies~\cite{niu2021hdr,prabhakar2021labeled} to handle ghosting artifacts, achieving state-of-the-art performance. Nonetheless, the restraints appear when confronted with long-range object movements and heavy intensity variations. Fig.~\ref{fig:teaser} shows a representative scene where large motions and severe saturations occur, producing unexpected ghosting and distortion artifacts in the results of previous CNN-based methods. The reason lies in the intrinsic locality restriction of convolution. CNN needs to stack deep layers to obtain a large receptive field and is thus ineffective to model long-range dependency (e.g., ghosting artifacts caused by large motion)~\cite{naseer2021intriguing}. Moreover, convolutions are content-independent as the same kernels are shared within the whole image, ignoring the long-range intensity variations of different image regions~\cite{liang2021swinir}. Therefore, exploring content-dependent algorithms with long-range modeling capability is demanding for further performance improvement.

Vision Transformer (ViT)~\cite{dosovitskiy2020image} has recently received increasing research interest due to its superior long-range modeling capability. However, our experimental results indicate two major issues that hinder its applications on HDR deghosting. On the one hand, Transformers lack the inductive biases inherent to CNN and therefore do not generalize well when trained on insufficient amounts of data~\cite{dosovitskiy2020image,liang2021swinir}, despite the fact that available datasets for HDR deghosting are limited as gathering huge numbers of realistic labeled samples is prohibitively expensive. On the other hand, the neighbor pixel relationships of both intra-frame and inter-frame are critical for recovering local details across multiple frames, while the pure Transformer is ineffective for extracting such local context.

To this end, we propose a novel Context-Aware Vision Transformer (CA-ViT), which is formulated to concurrently capture both global and local dependencies with a dual-branch architecture. For the global branch, we employ a window-based multi-head Transformer encoder to capture long-range contexts. For the local branch, we design a local context extractor (LCE), which extracts the local feature maps through a convolutional block and selects the most useful features across multiple frames by channel attention mechanism. The proposed CA-ViT, therefore, makes local and global contexts work in a complementary manner. By incorporating with the CA-ViT, we propose a novel Transformer-based framework (termed as HDR-Transformer) for ghost-free HDR imaging.

Specifically, the proposed HDR-Transformer mainly consists of a feature extraction network and an HDR reconstruction network. The feature extraction network extracts shallow features and fuses them coarsely through a spatial attention module. The early convolutional layers can stabilize the training process of the vision Transformer and the spatial attention module helps to suppress undesired misalignment. The HDR reconstruction network takes the proposed CA-ViT as basic components and is constituted hierarchically. The CA-ViTs model both long-range ghosting artifacts and local pixel relationship, thus helping to reconstruct ghost-free high-quality HDR images (an example is shown in Fig.~\ref{fig:teaser}) without the need of stacking very deep convolution blocks. In summary, the main contributions of this paper can be concluded as follows:

\begin{itemize}
  \item We propose a new vision Transformer, called CA-ViT, which can fully exploit both global and local image context dependencies, showing significant performance improvements over prior counterparts.
  \item We present a novel HDR-Transformer that is capable of removing ghosting artifacts and reconstructing high-quality HDR images with lower computational costs. To our best knowledge, this is the first Transformer-based framework for HDR deghosting.
  \item We conduct extensive experiments on three representative benchmark HDR datasets, which demonstrates the effectiveness of HDR-Transformer against existing state-of-the-art methods.
\end{itemize}


\section{Related Work}
\label{sec:related_work}

\subsection{HDR Deghosting Algorithms} 
We summarize existing HDR deghosting algorithms into three categories, i.e., motion rejection methods, image registration methods, and CNN-based methods.

\mysubtitle{Motion rejection methods} Methods based on motion rejection proposed first to register the LDR images globally and then reject the pixels which are detected as misaligned. Grosch \emph{et al.} generated an error map based on the alignment color differences to reject mismatched pixels~\cite{grosch2006fast}. Pece \emph{et al.} detected motion areas using a median threshold bitmap for input LDR images~\cite{pece2010bitmap}. Jacobs \emph{et al.} identified misaligned locations using weighted intensity variance analysis~\cite{jacobs2008automatic}. Zhang \emph{et al.}~\cite{zhang2011gradient} and Khan \emph{et al.}~\cite{khan2006ghost} proposed to calculate gradient-domain weight maps and probability maps for the LDR input images, respectively. Additionally, Oh \emph{et al.} presented a rank minimization method for the purpose of detecting ghosting regions~\cite{oh2014robust}. These methods frequently produce unpleasing HDR results due to the loss of useful information while rejecting pixels.

\mysubtitle{Motion registration methods} Motion registration methods rely on aligning the non-reference LDR images to the reference one before merging them. Begoni \emph{et al} proposed using optical flow to predict motion vectors~\cite{bogoni2000extending}. Kang \emph{et al.} transferred the LDR picture intensities to the luminance domain based on the exposure time and then estimated optical flow to account for motion~\cite{kang2003high}. Zimmer \emph{et al.} reconstructed the HDR image by first registering the LDR images with optical flow~\cite{zimmer2011freehand}. Sen \emph{et al.} presented a patch-based energy minimization method that simultaneously optimizes alignment and HDR reconstruction~\cite{sen2012robust}. Hu \emph{et al.} proposed to optimize the image alignment using brightness and gradient consistencies on the transformed domain~\cite{hu2013hdr}. Motion registration methods are more robust than motion rejection methods. However, when large motions occur, this approach generates visible ghosting artifacts.

\mysubtitle{CNN-based methods} Several CNN-based methods have been recently proposed. Kalantari \emph{et al.} proposed the first CNN-based method for multi-frame HDR imaging of dynamic scenes. They employed a CNN to blend the LDR images after aligning them with optical flow~\cite{kalantari2017deep}. Wu \emph{et al.} developed the first non-flow-based framework by formulating HDR imaging as an image translation problem~\cite{wu2018deep}. Instead of using explicit alignment, Yan \emph{et al.} adopted a spatial attention module to address ghosting artifacts~\cite{yan2019attention}. Prabhakar \emph{et al.} proposed an efficient method to generate HDR images with bilateral guided upsampler~\cite{prabhakar2020towards} and further explored zero and few-shot learning for HDR Deghosting~\cite{prabhakar2021labeled}. Lately, Niu \emph{et al.} proposed the first GAN-based framework for multi-frame HDR imaging~\cite{niu2021hdr}. The approaches based on CNNs demonstrate superior capabilities and achieve state-of-the-art performance. However, ghosting artifacts can still be observed when confronted with large motion and extreme saturation.

\subsection{Vision Transformers}
Transformers have achieved huge success in the field of natural language processing~\cite{vaswani2017attention,devlin2018bert}, where the multi-head self-attention mechanism is employed to capture long-range correlations between word token embeddings. Recently, ViT~\cite{dosovitskiy2020image} has shown that a pure Transformer can be applied directly to sequences of non-overlapping image patches and performs very well on image classification tasks. Liu \emph{et al.} developed Swin Transformer, a hierarchical structure where cross-window contexts are captured through the shift-window scheme~\cite{liu2021Swin}. Chen \emph{et al.} built IPT, a pretrained Transformer model for low-level computer vision tasks~\cite{chen2021pre}. Liang \emph{et al.} extended the Swin Transformer for image restoration and proposed SwinIR, achieving state-of-the-art performance on image super-resolution and denoising~\cite{liang2021swinir}. Unlike CNN-based methods, our approach is inspired by~\cite{liu2021Swin,liang2021swinir} and built on Transformers.

\section{Method}
\label{sec:method}

\begin{figure}[t]
  \centering
  \includegraphics[width=1.0\linewidth]{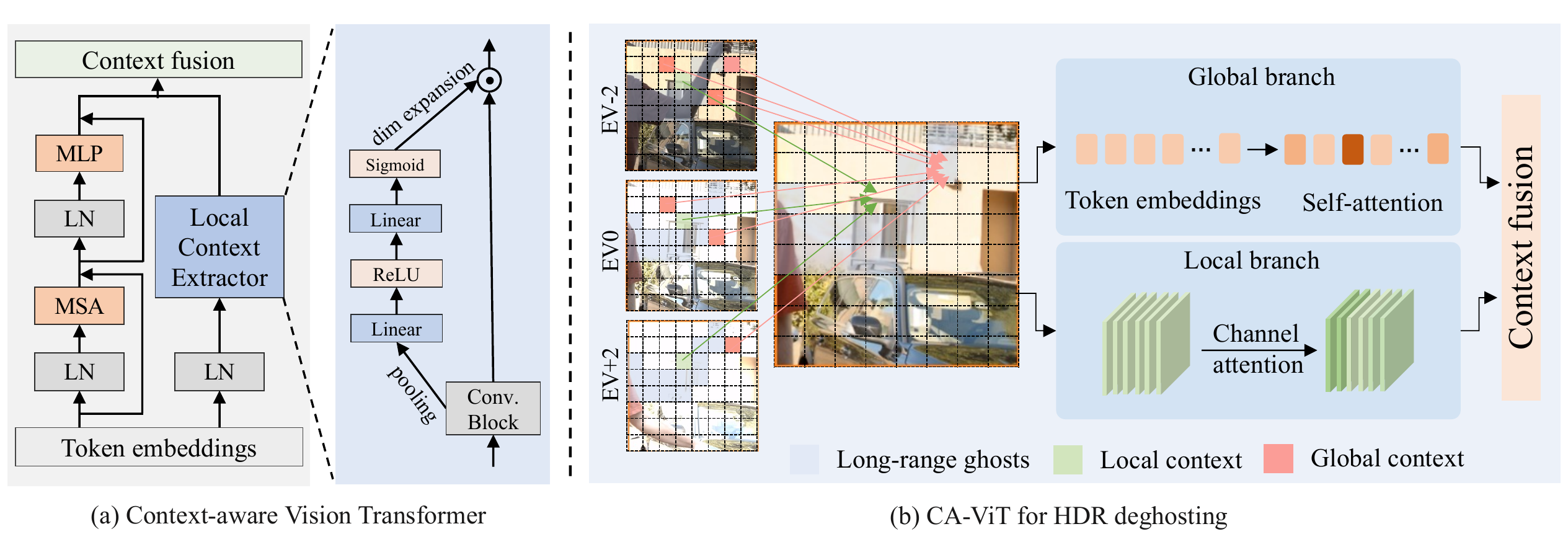}   
  \caption{Illustration of the proposed CA-ViT. As shown in Fig.~\ref{fig:ctl} (a), the CA-ViT is designed as a dual-branch architecture where the global branch models long-range dependency among image contexts through a multi-head Transformer encoder, and the local branch explores both intra-frame local details and inner-frame feature relationship through a local context extractor. Fig.~\ref{fig:ctl} (b) depicts the key insight of our HDR deghosting approach with CA-ViT. To remove the residual ghosting artifacts caused by large motions of the hand (marked with blue), long-range contexts (marked with red), which are required to hallucinate reasonable content in the ghosting area, are modeled by the self-attention in the global branch. Meanwhile, the well-exposed non-occluded local regions (marked with green) can be effectively extracted with convolutional layers and fused by the channel attention in the local branch.}\label{fig:ctl}
\end{figure}

\subsection{CA-ViT}
\label{sec:ctl}
Unlike prior vision Transformers that adopt the pure Transformer encoder, we propose a dual-branch context-aware vision Transformer (CA-ViT), which explores both the global and local image information. As depicted in Fig.~\ref{fig:ctl} (a), the proposed CA-ViT is constructed with a global Transformer encoder branch and a local context extractor branch.

\subsubsection{Global Transformer Encoder}
For the global branch, we employ a window-based multi-head Transformer encoder~\cite{dosovitskiy2020image} to capture long-range information. The Transformer encoder consists of a multi-head self-attention (MSA) module and a multi-layer perceptron (MLP) with residual connection.

Considering the input token embeddings $E\in \mathbb{R}^{H \times W \times D}$, the global context branch can be formulated as:
\begin{equation}
  \begin{aligned}\label{eq:ctl_global}
    E = \textit{MSA}({\textit{LN}(E)}) + E, \\
    \textit{CTX}_{global} = \textit{MLP}({\textit{LN}(E)}) + E,
  \end{aligned}
\end{equation}
where $\textit{LN}$ denotes LayerNorm, and $\textit{CTX}_{global}$ denotes the global contexts captured by the Transformer encoder.

\subsubsection{Local Feature Extractor}
For the local branch, we design a local context extractor (LCE) to extract local information $\textit{CTX}_{local}$ from adjacent pixels and select cross-channel features for fusion, which is defined as:
\begin{align}\label{eq:ctl_local}
  \textit{CTX}_{local} = \textit{LCE}({\textit{LN}(E)}).
\end{align}

Specifically, for the token embeddings $\textit{E}$ normalized with an LN layer, we first reshape them into $H \times W\times D$ features and use a convolution block to extract local feature maps $f_{local}$. The local features are then average pooled to a shape of $1 \times 1\times D$, and the channel-wise weights $\omega$ are calculated from two linear layers followed by a ReLU and a sigmoid activation layer, respectively. Afterward, the useful feature maps are selected through a channel-wise calibration from the original local features $f_{local}$, i.e.,
\begin{equation}
  \begin{aligned}\label{eq:lce}
  f_{local} &= \textit{Conv}(\textit{LN}(E)),\\
  \omega &= \sigma_2(\textit{FC}(\sigma_1(\textit{FC}(f_{local})))), \\
  \textit{CTX}_{local} &= \omega \odot f_{local}, \\
  \end{aligned}
\end{equation}
where $\sigma_1$ and $\sigma_2$ denote the ReLU and sigmoid layer, and $\textit{FC}$ denotes the linear layer. As a result, the local context branch not only adds the locality into the Transformer encoder, but also identifies the most informative local features across multiple frames for feature fusion.

Finally, a context fusion layer is employed to combine the global and local contexts.  Although other transformation functions (e.g., linear or convolution layer) can be used to implement the context fusion layer, in this paper, we simply merge the contexts by element-wise addition to reduce the influence of additional parameters.

\subsection{HDR Deghosting}
The task of deep HDR deghosting aims to reconstruct a ghost-free HDR image through deep neural networks. Following most of the previous works~\cite{kalantari2017deep,wu2018deep,yan2019attention}, we consider 3 LDR images (i.e., $I_i, i=1,2,3$) as input and refer to the middle frame $I_2$ as the reference image. To better utilize the input data, the LDR images $\{I_i\}$ are first mapped to the HDR domain using the gamma correction, generating the gamma-corrected images $\{\check{I}_i\}$: 
\begin{align}\label{eq:gamma_correction}
  \quad \check{I}_{i} = \frac{(I_{i})^\gamma}{t_{i}}, \quad i=1, 2, 3,
\end{align}
where $t_i$ denotes the exposure time of $I_i$, and $\gamma$ is the gamma correction parameter, which is set to 2.2 in this paper. We then concatenate the original LDR images $\{I_i\}$ and the corresponding gamma-corrected images $\{\check{I}_i\}$ into a 6-channels input $\{X_i\}$. This strategy is suggested in~\cite{kalantari2017deep} as the LDR images help to detect the noisy or saturated regions, while the gamma-corrected images are helpful for detecting misalignments. Finally, the network $\varPhi(\cdot)$ is defined as:
\begin{align}\label{eq:network}
  I^{\mathrm{\hat{H}}} = \varPhi (X_i; \theta), \quad i=1, 2, 3,
\end{align}
where $I^{\mathrm{\hat{H}}}$ denotes the reconstructed HDR image, and $\theta$ is the network parameters to be optimized.

Instead of stacking very deep CNN layers to obtain a large receptive field as existing CNN-based approaches, we propose the HDR-Transformer to handle HDR deghosting. Our key insight is that, with the specifically-designed dual-branch CA-ViT, the long-range ghosting can be well modeled in the global branch, and the local branch helps to recover fine-grained details. We describe the architecture of the proposed HDR-Transformer in the next section.

\begin{figure*}[t]
  \centering
  \includegraphics[width=1.0\linewidth]{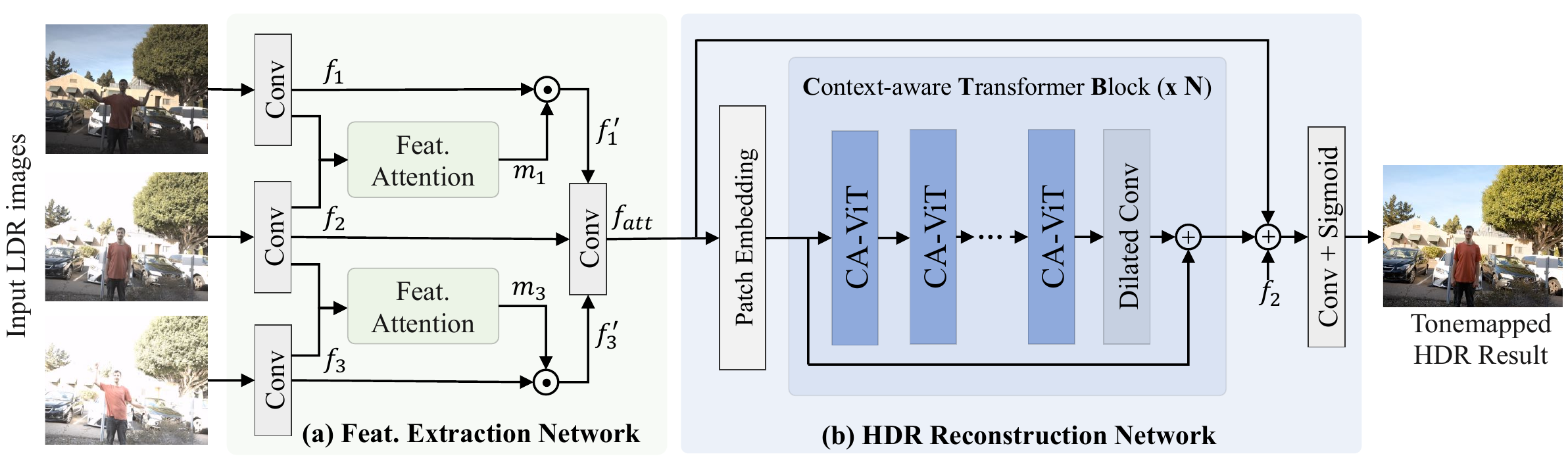}   
  \caption{The network architecture of HDR-Transformer. The pipeline consists of two stages: (a) The feature extraction network first extracts the coarse features through a spatial attention module. (b) The extracted features are then fed into the HDR reconstruction network to recover the HDR results. The HDR reconstruction network consists of several Context-aware Transformer Blocks (CTBs), which take the proposed CA-ViT as basic components.}\label{fig:pipeline}
\end{figure*}

\subsection{Overall Architecture of HDR-Transformer}
As illustrated in Fig.~\ref{fig:pipeline}, the overall structure of our proposed HDR-Transformer mainly consists of two components, i.e., feature extraction network (Fig.~\ref{fig:pipeline} (a)) and HDR reconstruction network (Fig.~\ref{fig:pipeline} (b)). Given three input images, we first extract the spatial features through a spatial attention module. The extracted coarser features are then embedded and fed into the Transformer-based HDR reconstruction network, generating the reconstructed ghost-free HDR image.

\subsubsection{Feature Extraction Network} 
The early convolution layers help to stabilize the training process of Vision Transformers~\cite{xiao2021early}. For the input images $X_{i} \in \mathbb{R}^{H \times W \times 6}, i=1, 2, 3$, we first extract the shallow features $f_{i} \in \mathbb{R}^{H \times W \times C}$ by three separate convolution layers, where $C$ is the number of channels. Then, we concatenate each non-reference feature (i.e., $f_{1}$ and $f_{3}$) with the reference feature $f_{2}$ and calculate the attention maps $m_{i}$ through a spatial attention module $\mathcal{A}$:
\begin{align}\label{eq:att_maps}
  m_{i} = \mathcal{A} (f_{i}, f_{2}), \quad i=1, 3,
\end{align}
The attention features $f^{'}_{i}$ are computed by multiplying the attention maps $m_i$ by the non-reference features $f_i$, i.e.,
\begin{align}\label{eq:att_feats}
  f^{'}_{i} = f_{i} \odot  m_{i}, \quad i=1, 3,
\end{align}
where $\odot$ denotes the element-wise multiplication. The spatial attention module has been proved to effectively reduce undesired contents caused by foreground object movements~\cite{yan2019attention,liu2021adnet}. The convolution layers in the attention module can also increase the inductive biases for the subsequent Transformer layers. 

\subsubsection{HDR Reconstruction Network} 
As shown in Fig.~\ref{fig:pipeline}, the HDR reconstruction network is mainly composed of several context-aware Transformer blocks (CTBs). The input of the first CTB $f_{att} \in \mathbb{R}^{H \times W \times D}$ is obtained from $f^{'}_{1}$, $f_{2}$, and $f^{'}_{3}$ and embedded into token embeddings, where $\textit{D}$ denotes the embed dimension. The HDR result is reconstructed by $\textit{N}$ subsequent CTBs and a following convolution block. We also adopt the global skip connection to stabilize the optimization process.

\mysubtitle{Context-aware Transformer Block}
As illustrated in Fig.~\ref{fig:ctl} (b), when suffering occlusion caused by large object movements and heavy saturation, long-range context is required for removing the corresponding ghosting regions and hallucinating reasonable content, while the non-occluded areas can be fused well by the convolutional layers. To this end, we develop the context-aware Transformer block (CTB) by taking the proposed CA-ViT as the basic component.

For clarity, each CTB contains $\textit{M}$ CA-ViTs. For the $n$-th CTB with the input of $F_{n,0}$, the output of the $m$-th CA-ViT can be formulated as:
\begin{align}\label{eq:cstb_ctl}
  F_{n,m} =  \mathcal{C}_{n,m}(F_{n,m-1}), \quad m=1, 2,..., M,
\end{align}
where $C_{n, m}(\cdot)$ denotes the corresponding CA-ViT. Then, we feed the output of the $\textit{M}$-th CA-ViT into a dilated convolution layer. The dilated convolutional layer is employed to increase the receptive field of the context range.  We also adopt the residual connection in each CTB for better convergence. Consequently, the output of the $n$-th CTB is formulated as:
\begin{align}\label{eq:cstb_out}
  F_{n} =  \textit{DConv}(F_{n,M}) + F_{n, 0},
\end{align}
where $DConv(\cdot)$ denotes the dilated convolutional layer, and $\textit{M}$ and $\textit{N}$ are empirically set to 6 and 3, respectively.

\subsection{Loss Function}
As HDR images are typically viewed after tonemapping, we compute the loss in the tonemapped domain using the commonly used $\mu$-law function:
\begin{align}\label{eq:mu_law}
  \mathcal{T}(x)=\frac{\log (1+\mu x)}{\log(1+\mu)},
\end{align}
where $\mathcal{T}(x)$ is the tonemapped HDR image, and we set $\mu$ to 5000. Unlike previous methods~\cite{kalantari2017deep,wu2018deep,yan2019attention} that only adopt the pixel-wise loss (e.g., $l_1$ or $l_2$ error), we utilize $l_1$ loss and perceptual loss to optimize the proposed HDR-Transformer. Given the estimated HDR image $I^{\hat{H}}$ and the ground truth HDR image $I^{H}$, the $l_1$ loss term is defined as:
\begin{align}\label{eq:recon_loss}
  \mathcal{L}_{r} = \parallel \mathcal{T}(I^{H}) - \mathcal{T}(I^{\hat{H}}) \parallel_{1},
\end{align}
The perceptual loss~\cite{Johnson2016Perceptual} is widely used in image inpainting~\cite{liu2018image} for better visual quality improvements. We also apply the perceptual loss to enhance the quality of the reconstructed HDR images:
\begin{align}\label{eq:percep_loss}
  \mathcal{L}_{p} = \sum_{j} \parallel \varPsi_{j}(\mathcal{T}(I^{H})) - \varPsi_{j}(\mathcal{T}(I^{\hat{H}})) \parallel_{1},
\end{align}
where $\varPsi(\cdot)$ denotes the activation feature maps extracted from a pre-trained VGG-16 network~\cite{simonyan2014very}, and $j$ denotes the $j$-th layer. We analyze the effectiveness of the perceptual loss in our ablation study (Sec.~\ref{sec:ablation_loss}). Eventually, our training loss function $\mathcal{L}$ is formulated as:
\begin{align}\label{eq:loss}
  \mathcal{L} = \mathcal{L}_{r} + \lambda_{p}\mathcal{L}_{p},
\end{align}
where $\lambda_{p}$ is the hyper-parameter and we set it to 0.01.

\section{Experiments}
\label{sec:experiments}

\subsection{Dataset and Implementation Details}
\mysubtitle{Datasets} Following previous methods~\cite{wu2018deep,yan2019attention,yan2020deep,niu2021hdr}, we train our network on the widely used Kalantari \emph{et al.}'s dataset~\cite{kalantari2017deep}, which consists of 74 samples for training and 15 samples for testing. Each sample from Kalantari \emph{et al.}'s dataset comprises three LDR images with exposure values of $\left\langle-2, 0, +2\right\rangle$ or $\left\langle-3, 0, +3\right\rangle$, as well as a ground truth HDR image. During the training, we first crop patches of size $128\times128$ with a stride of 64 from the training set. We then apply rotation and flipping augmentation to increase the training size. We quantitatively and qualitatively evaluate our method on Kalantari \emph{et al.}'s testing set. We also conduct evaluations on Sen \emph{et al.}~\cite{sen2012robust}'s and Tursun \emph{et al.}~\cite{tursun2016objective}'s datasets to verify the generalization ability of our method.

\mysubtitle{Evaluation Metrics} We use PSNR and SSIM as evaluation metrics. To be more precise, we calculate PSNR-$l$, PSNR-$\mu$, SSIM-$l$, and SSIM-$\mu$ scores between the reconstructed HDR images and their corresponding ground truth. The `-$l$' and `-$\mu$' denote the linear and tonemapped domain values, respectively. Given that HDR images are typically displayed on LDR displays, metrics in the tonemapped domain more accurately reflect the quality of the reconstructed HDR images. Additionally, we conduct evaluations using the HDR-VDP-2~\cite{mantiuk2011hdr}, which is developed specifically for evaluating the quality of HDR images.

\mysubtitle{Implementation Details} Our HDR-Transformer is implemented by PyTorch. We use the ADAM optimizer with an initial learning rate of 2e-4 and set $\beta_{1}$ to 0.9, $\beta_{2}$ to 0.999, and $\epsilon$ to 1e-8, respectively. We train the network from scratch with a batch size of 16 and 100 epochs enables it to converge. The whole training is conducted on four NVIDIA 2080Ti GPUs and costs about two days.

\begin{table}[t]
  \centering
  \caption{Quantitative comparison between previous methods and ours on Kalantari \emph{et al.}~\cite{kalantari2017deep}'s test set. We use PSNR, SSIM, and HDR-VDP-2 as evaluation metrics. The `-$\mu$' and `-$l$' refers to values calculated on the tonemapped domain and the linear domain, respectively. All values are the average over 15 testing images and higher better. The best results are highlighted and the second best are underlined.}
  \label{tab:results}
  \resizebox{1.0\linewidth}{!}{
    \begin{tabular}{
    c
    >{\centering\arraybackslash}p{1.4cm}
    >{\centering\arraybackslash}p{1.4cm}
    >{\centering\arraybackslash}p{1.6cm}
    >{\centering\arraybackslash}p{1.6cm}
    >{\centering\arraybackslash}p{1.6cm}
    >{\centering\arraybackslash}p{1.6cm}
    >{\centering\arraybackslash}p{1.8cm}
    >{\centering\arraybackslash}p{1.4cm}
    >{\centering\arraybackslash}p{2.6cm}}
  \toprule
  \multirow{3}{*}{Metrics} & \multicolumn{8}{c}{Methods}                                                     \\ \cmidrule(l){2-10} 
              & Sen12  & Hu13   & Kalantari17 & DeepHDR & AHDRNet & NHDRRNet & HDR-GAN & SwinIR & HDR-Transformer  \\
              & \cite{sen2012robust}  & \cite{hu2013hdr} & \cite{kalantari2017deep} & \cite{wu2018deep} & \cite{yan2019attention}  & \cite{yan2020deep}  & \cite{niu2021hdr} &\cite{liang2021swinir} & Ours\\  \midrule
  PSNR-$\mu$     & 40.80    & 35.79   &  42.67   &  41.65   &  43.63  & 42.41  & \underline{43.92} & 43.42 & \textbf{44.32}  \\
  PNRR-$l$      & 38.11    & 30.76   &  41.23   &   40.88  &  41.14  & 41.43  & 41.57  & \underline{41.68} & \textbf{42.18}  \\
  SSIM-$\mu$      & 0.9808    & 0.9717   &  0.9888  &  0.9860  &  0.9900 & 0.9877 & \underline{0.9905} & 0.9882 & \textbf{0.9916}  \\
  SSIM-$l$      & 0.9721    & 0.9503   &  0.9846  &  0.9858  &  0.9702  & 0.9857 & \underline{0.9865}  & 0.9861 & \textbf{0.9884} \\
  HDR-VDP-2 & 59.38    &  57.05  &   65.05     & 64.90  &  64.61 & 61.21 & \underline{65.45}  & 64.52 & \textbf{66.03} \\ \bottomrule
  \end{tabular}
  }
\end{table}

\begin{figure*}[t]
  \centering
  \includegraphics[width=1.0\linewidth]{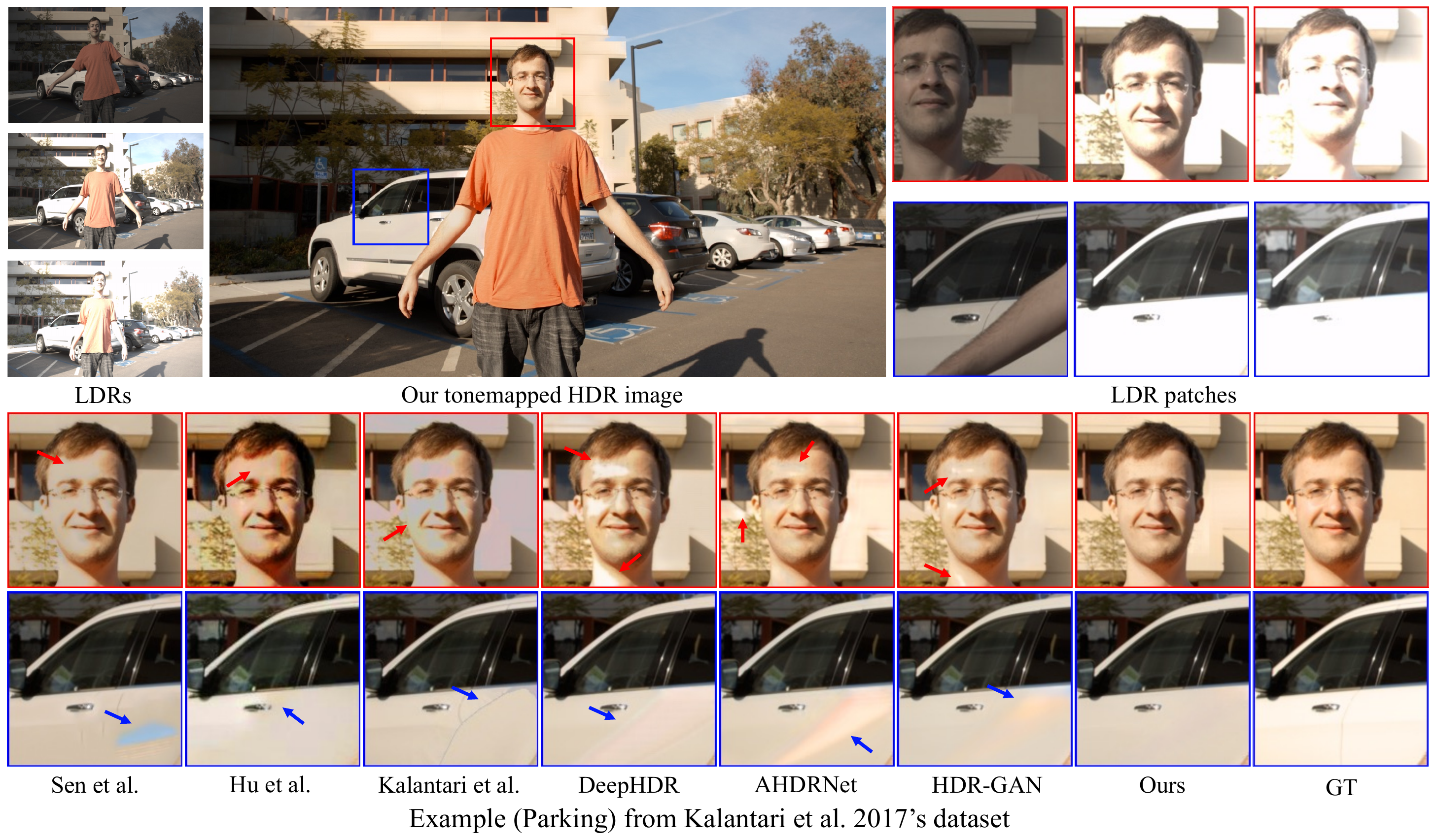}   
  \caption{More visual comparisons between the proposed method and state-of-the-art methods~\cite{sen2012robust,hu2013hdr,kalantari2017deep,wu2018deep,yan2019attention,niu2021hdr} on Kalantari \emph{et al.}~\cite{kalantari2017deep}’s dataset.}\label{fig:result_1}
\end{figure*}

\subsection{Comparison with State-of-the-art Methods}

\subsubsection{Results on Kalantari~\emph{et al.}'s Dataset}
We first compare the results of the proposed HDR-Transformer with several state-of-the-art methods, which include two patch match based methods (Sen \emph{et al.}~\cite{sen2012robust} and Hu \emph{et al.}~\cite{hu2013hdr}) and five CNN-based methods (Kalantari \emph{et al.}~\cite{kalantari2017deep}, DeepHDR~\cite{wu2018deep}, AHDRNet~\cite{yan2019attention}, NHDRRNet~\cite{yan2020deep}, and HDR-GAN~\cite{niu2021hdr}). We also compare with a tiny version of SwinIR~\cite{liang2021swinir} as the original one fails to converge on the limited dataset. Among the deep learning-based methods, Kalantari \emph{et al.}~\cite{kalantari2017deep} adopt optical flow to align the input LDR images while DeepHDR~\cite{wu2018deep} aligns the background using homography. In contrast, the left approaches and our HDR-Transformer don't require any pre-alignment. We report the quantitative and qualitative comparison results as this testing set contains ground truth HDR images.

\mysubtitle{Quantitative results} Table~\ref{tab:results} lists the quantitative results. For the sake of fairness, the results of prior works are borrowed from HDR-GAN~\cite{niu2021hdr}, and all results are averaged over 15 testing samples from Kalantari \emph{et al.}'s dataset. Several conclusions can be drawn from Table~\ref{tab:results}. Firstly, all deep learning-based algorithms have demonstrated significant performance advantages over patch match based methods. Secondly, the pure Transformer encoder adopted in SwinIR doesn't perform well for the aforementioned reasons. Thirdly, the proposed HDR-Transformer surpasses the recently published HDR-GAN~\cite{niu2021hdr} by up to 0.6dB and 0.4dB in terms of PSNR-$l$ and PSNR-$\mu$, respectively, demonstrating the effectiveness of our method.

\mysubtitle{Qualitative results} For fair comparisons, all qualitative results are obtained using the codes provided by the authors and tonemapped using the same settings in Photomatix Pro. Fig.~\ref{fig:result_1} illustrates an intractable scene that contains saturations and large motion. The first row shows the input LDR images, our tonemapped HDR result, and the corresponding zoomed LDR patches from left to right. The second row lists the compared HDR results, where the two comparison locations are highlighted in red and blue, respectively. As can be seen, the red boxed area suffers heavy intensity variation within the three input LDR images and causes long-range saturation. Previous approaches remove the ghosting artifacts induced by slight head movements but fail to hallucinate the details of the saturation regions on the face, resulting in color distortions and inconsistent details. The blue boxed patches show a large motion region caused by the hand, patch match based methods fail to discover the correct regions, and CNN-based methods fail to handle the long-range motion, leading to ghosting artifacts in the reconstructed HDR image. On the contrary, The proposed HDR-Transformer reconstructs ghost-free results while hallucinating more visually pleasing details in these areas.

\begin{figure}[t]
  \centering
  \includegraphics[width=1.0\linewidth]{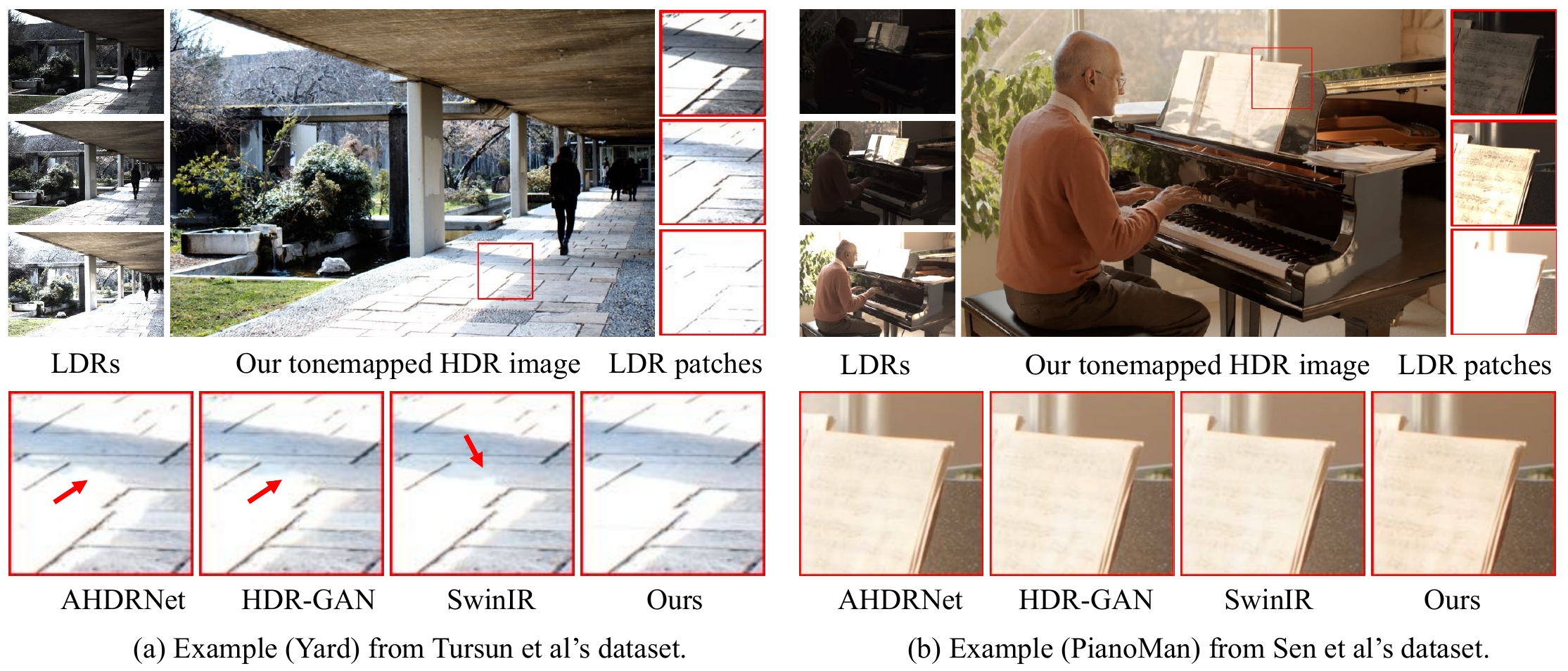}   
  \caption{Comparison results on the datasets without ground truth. Scenes are obtained from the Tursun \emph{et al.}~\cite{tursun2016objective}'s and the Sen \emph{et al.}~\cite{sen2012robust}'s datasets. Our approach generates better results in the saturated boundary and hallucinates more high-frequency details when suffering heavy intensity variation.}\label{fig:result_sen_tursun}
\end{figure}

\subsubsection{Results on the Datasets w/o Ground Truth}
To validate the generalization ability of our method, we conduct evaluations on Sen \emph{et al.}~\cite{sen2012robust}'s and Tursun \emph{et al.}~\cite{tursun2016objective}'s datasets. As illustrated in Fig.~\ref{fig:result_sen_tursun}, we report the qualitative results as both datasets have no ground truth HDR images. As seen in Fig.~\ref{fig:result_sen_tursun} (a), When suffering long-range saturation, the CNN-based algorithms AHDRNet~\cite{yan2019attention} and HDR-GAN~\cite{niu2021hdr} produce undesired distortions in saturated boundaries. The Transformer-based method SwinIR~\cite{liang2021swinir} performs better but still contains noticeable distortion as the inefficiency of local context modeling. On the contrary, the proposed HDR-Transformer generates more precise boundaries (best to compare with the corresponding LDR patches), demonstrating the context-aware modeling ability of our method. Fig.~\ref{fig:result_sen_tursun} (b) shows a scene where the piano spectrum gets saturated. Previous methods lose the high-frequency details and produce blurry results, while our approach hallucinates more details than them.

\subsubsection{Analysis of Computational Budgets}
We also compare the inference times and model parameters with previous works. As shown in Table~\ref{tab:running_time}, the patch match based methods~\cite{sen2012robust,hu2013hdr} take more than 60 seconds to fuse a 1.5MP LDR sequence. Among the CNN-based methods, Kalantari \emph{et al.}~\cite{kalantari2017deep} costs more time than the left non-flow based methods because of the time-consuming optical flow preprocess. DeepHDR~\cite{wu2018deep} and NHDRRNet~\cite{yan2020deep} consume fewer inference times but need huge amounts of parameters. AHDRNet~\cite{yan2019attention} and HDR-GAN~\cite{niu2021hdr} have a better balance of performance and efficiency by taking advantage of their well-designed architectures. In contrast, HDR-Transformer outperforms the state-of-the-art method HDR-GAN~\cite{niu2021hdr} with only half computational budgets.

\begin{table}[t]
  \centering
  \caption{The inference times and parameters of different methods. Part of the values are from~\cite{yan2020deep}. The `-' denotes the patch match based methods have no parameters.}
  \label{tab:running_time}
  \resizebox{1.0\linewidth}{!}{
    \begin{tabular}{
    c
    >{\centering\arraybackslash}p{1.0cm}
    >{\centering\arraybackslash}p{1.0cm}
    >{\centering\arraybackslash}p{1.6cm}
    >{\centering\arraybackslash}p{1.6cm}
    >{\centering\arraybackslash}p{1.6cm}
    >{\centering\arraybackslash}p{1.6cm}
    >{\centering\arraybackslash}p{1.8cm}
    >{\centering\arraybackslash}p{2.8cm}}
  \toprule
  \multirow{2}{*}{Method} & Sen12 & Hu13 & Kalantari17 & DeepHDR & AHDRNet & NHDRRNet & HDR-GAN & HDR-Transformer \\ 
                        &  \cite{sen2012robust} & \cite{hu2013hdr}   & \cite{kalantari2017deep}   & \cite{wu2018deep}&\cite{yan2019attention}&\cite{yan2020deep}& \cite{niu2021hdr}& Ours \\ \midrule
  Environment   & CPU    & CPU   &  CPU+GPU   &  GPU  &  GPU  & GPU  & GPU &  GPU \\ 
  Time(s)   & 61.81s    & 79.77s   &  29.14s   &   0.24s  &  0.30s  & 0.31s  & 0.29s  &  0.15s  \\
  Parameters(M) & -   &  -  & 0.3M  & 20.4M  &  1.24M & 38.1M & 2.56M  & 1.22M \\ \bottomrule
  \end{tabular}
  }
\end{table}

\subsection{Ablation Study}
\label{sec:balation_study}
To analyze the effectiveness of each component, we conduct comprehensive ablation studies on Kalantari \emph{et al.}~\cite{kalantari2017deep}'s dataset. We report the PSNR and HDR-VDP-2 scores for quantitative comparison. 

\subsubsection{Ablation on the network architecture}
For the network design, we compare the proposed CA-ViT, the adopted spatial attention (SA) module, and the overall HDR-Transformer with the baseline model. Specifically, we design the following variants: 

\begin{itemize}
  \item \textbf{Baseline}. We take a tiny version of SwinIR~\cite{liang2021swinir}, which is constituted with vanilla Transformer encoders, as our baseline model. The baseline model keeps comparable network parameters and the same training settings as our proposed HDR-Transformer.
  \item +\textbf{\ CA-ViT}. This variant replaces the vanilla Transformer encoder used in the baseline model with the proposed Context-aware Vision Transformer.
  \item +\textbf{\ SA}. In this variant, we add a spatial attention (SA) module to fuse the shallow features extracted from the three input LDR images.
  \item +\textbf{\ CA-ViT\ }+\textbf{\ SA}. The overall network of the proposed HDR-Transformer.
\end{itemize}

\begin{table}[t]
  \caption{Quantitative results of the ablation studies. BL: the baseline model, CA-ViT: the proposed Context-aware Vision Transformer, SA: the spatial attention module, $\mathcal{L}_p$: the perceptual loss term.}
  \label{tab:ablation_study}
  \centering
  \resizebox{0.6\linewidth}{!}{

    \begin{tabular}{
    >{\centering\arraybackslash}p{0.85cm}
    >{\centering\arraybackslash}p{1.2cm}
    >{\centering\arraybackslash}p{0.85cm}
    >{\centering\arraybackslash}p{0.85cm}
    >{\centering\arraybackslash}p{1.6cm}
    >{\centering\arraybackslash}p{1.6cm}
    >{\centering\arraybackslash}p{2.0cm}}
    \toprule
    BL & CA-ViT & SA & $\mathcal{L}_p$ & PSNR-$\mu$ & PSNR-$l$ & HDR-VDP-2 \\ \midrule
    \checkmark  &     &       &    &   43.42   &  41.68   &   64.52     \\
    \checkmark  & \checkmark  &   &    &  44.03    &  41.99    &  65.94      \\
    \checkmark  &  & \checkmark    &    &  43.77    &  41.78    &  65.30      \\
    \checkmark  & \checkmark   & \checkmark     &   & 44.26   & 42.09  &   65.97     \\
    \checkmark  & \checkmark   & \checkmark     & \checkmark  &  44.32    &  42.18   &   66.03   \\ \bottomrule   
    \end{tabular}
  }
  \end{table}

 \begin{figure}[t]
  \centering
  \includegraphics[width=0.8\linewidth]{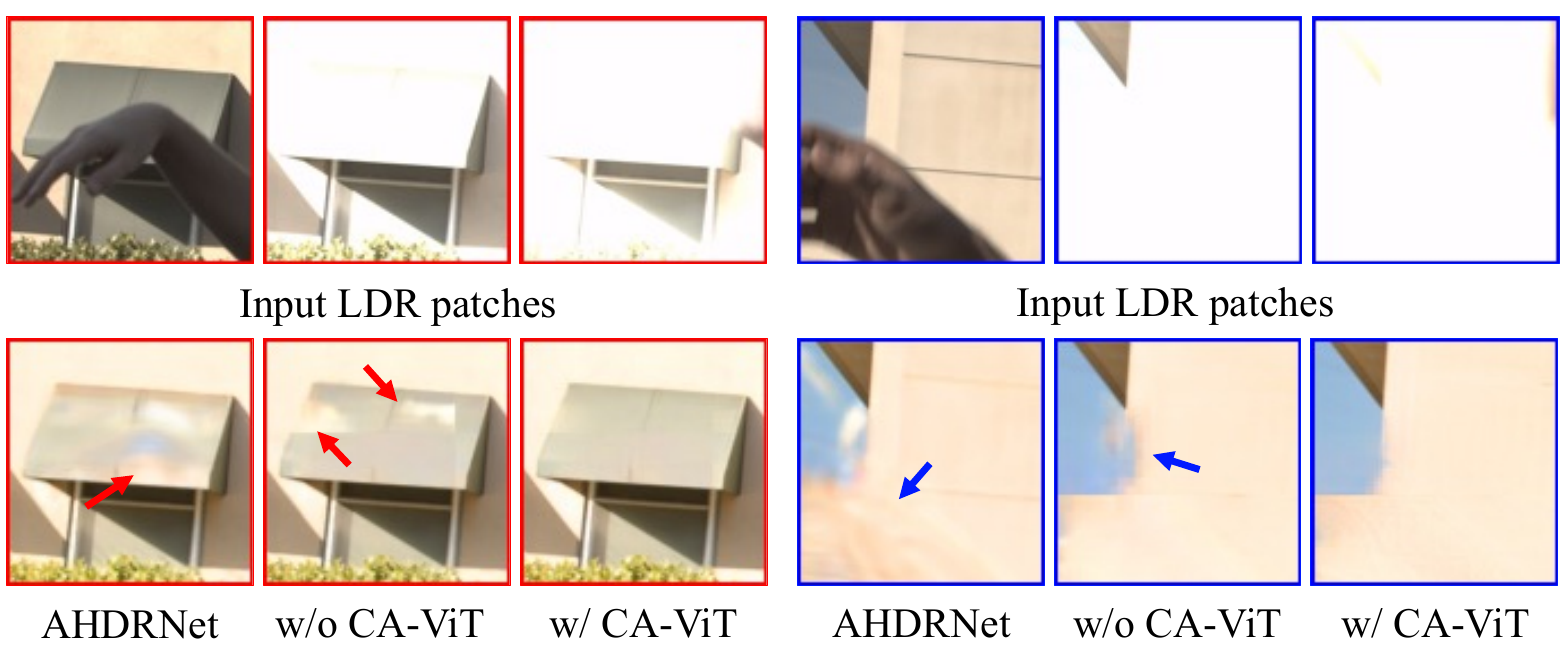}   
  \caption{Qualitative results of our ablation study on the proposed CA-ViT.}\label{fig:ablation_ctl}
\end{figure}

Table~\ref{tab:ablation_study} summarizes the quantitative results of our ablation study. The first row in Table~\ref{tab:ablation_study} shows that directly applying the Transformer to HDR deghosting does not perform well. By comparing the first four rows, several conclusions can be drawn. On the one hand, the CA-ViT and SA both improve the performance, but the benefit from CA-ViT is more significant than SA. We conclude the reasons in two folds. Firstly, the inductive biases introduced by the convolution layers in the CA-ViT or SA help the Transformer be better optimized in limited data. Moreover, by incorporating the CA-ViT into each Transformer encoder, both the global and local contexts are explored, resulting in better capabilities of long-range ghosting removal and local details reconstruction. The qualitative results in Fig.~\ref{fig:ablation_ctl} also demonstrate our conclusions. On the other hand, the performance is further improved by combining all the components, which proves the effectiveness of the HDR-Transformer's pipeline design.

\subsubsection{Ablation on losses} 
\label{sec:ablation_loss}

We also conduct experiments to verify the effectiveness of the perceptual loss by training the HDR-Transformer from scratch both with and without the perceptual loss term. Comparing the last two rows in Table~\ref{tab:ablation_study}, we can see that the adopted perceptual loss improves the performance of the proposed HDR-Transformer. 

\section{Conclusions}
\label{sec:conclusion}
In this paper, we have proposed a dual-branch Context-aware Vision Transformer (CA-ViT), which overcomes the lack of locality in vanilla ViTs. We have extended the standard ViTs by incorporating a local feature extractor, and therefore both global and local image contexts are modeled concurrently. Furthermore, we have introduced the HDR-Transformer, a task-specific framework for ghost-free high dynamic range imaging. The HDR-Transformer incorporates the benefits of Transformers and CNNs, where the Transformer encoder and the local context extractor are used to model the long-range ghosting artifacts and short-range pixel relationship, respectively. Extensive experiments have demonstrated that the proposed method achieves state-of-the-art performance.

\noindent\textbf{Acknowledgement} This work was supported by National Natural Science Foundation of China under grants No. (61872067, 62031009 and 61720106004).



\clearpage
%
%
\bibliographystyle{splncs04}
\bibliography{egbib}
\end{document}